\pdfoutput=1

\documentclass[english,runningheads]{lni}

\usepackage{tabularx}
\usepackage{csquotes}

\definecolor{Sepia}{RGB}{125,78,40}
\definecolor{OliveGreen}{RGB}{45,67,7}
\definecolor{RedViolet}{RGB}{117, 33, 98}
\definecolor{MidnightBlue}{RGB}{25,25,112}

\begin{document}

\title[Structured Documents by ChatGPT]{ Large Language Models are Pattern Matchers: Editing Semi-Structured and Structured Documents with ChatGPT }
\subtitle{Author's own version. To appear in  M. Böhm \& J. Wunderlich (Eds.), AKWI Jahrestagung 2024. DOI 10.18420/AKWI2024-001
} 
\author[1]{Irene Weber}{irene.weber@hs-kempten.de}{0000-0003-2743-1698}
\affil[1]{Kempten University of Applied Sciences\\Faculty of Mechanical Engineering  \\Bahnhofstraße 61\\87435 Kempten\\Germany}
\booktitle{ M. Böhm \& J. Wunderlich (Eds.), AKWI Jahrestagung 2024.}
\maketitle

\begin{abstract}
Large Language Models (LLMs) offer numerous applications, the full extent of which is not yet  understood. This paper investigates if  LLMs can be applied for editing structured and semi-structured documents with minimal effort. Using  a qualitative research approach, we conduct two case studies with ChatGPT and thoroughly analyze the  results.  Our experiments indicate that LLMs can effectively edit structured and semi-structured documents when provided with basic,  straightforward  prompts. ChatGPT demonstrates a strong ability to recognize and process the structure of  annotated documents. This suggests that explicitly structuring tasks and data in prompts might enhance an LLM's ability to understand and solve tasks. Furthermore, the experiments also reveal impressive pattern matching skills in ChatGPT.  This observation  deserves further investigation,  as it may contribute to understanding the processes leading to hallucinations in LLMs.

\end{abstract}

\section{Introduction}
Large Language Models (LLMs) are extensive artificial neural networks trained on vast amounts of textual data to generate coherent continuations of given prompts. The initial training, which is time-consuming and computationally intensive, is typically followed by additional training phases. Fine-tuning with specific tasks and example responses enables LLMs to solve particular types of problems, while Reinforcement Learning with Human Feedback focuses them on delivering high-quality and socially preferred responses. Research has shown that LLMs can not only produce correct natural and formal language texts conveying plausible contents, but are also capable of reasoning, planning, and simulating other forms of intelligent behaviors. Thus, LLMs offer a wide range of potential applications, the extent of which is still not fully explored. 

Frequently, LLMs are applied for creating and processing texts, for communicating, planning, and computer programming. LLMs require that  all tasks and inputs are provided in a textual format.  For many applications, LLMs are prompted with freely phrased, natural language text or program code. Yet, they are also capable of processing texts that are structured such that they represent data or formatted documents. 

The term \textit{unstructured document} refers to textually encoded information lacking explicit organization, such as natural-language text without a defined context or fixed format. \textit{Structured data} refers to information with an explicit and strict regular structure, like data originating from database management systems. In structured data, the meaning of a data element is defined by the structure in which it is registered, and the order of data elements, in general, is not meaningful. 
\textit{Semi-structured documents} fall between unstructured documents and structured data. They have a flexible structure, often combining heterogeneous textual contents, such as short, potentially ungrammatical text fragments, longer free-text, and markup tags. \cite{madaniSemistructuredDocumentsMining2013}

There are various methods for indicating structure in texts, including markup languages like Markdown or HTML, data exchange formats like XML, JSON, and YAML, formalized languages, and  tabular formats as e.g., comma-separated value (CSV) data. Specialized formalisms often build upon generic formats like XML or JSON, such as formalisms for representing process models or other types of graphs. 
Documents containing formatting markup, such as HTML or LaTeX, are generally considered semi-structured \cite{madaniSemistructuredDocumentsMining2013}.
XML, JSON, and similar formalisms can represent semi-structured documents as well as structured data, depending on the presence and flexibility of an underlying schema.
For the remainder of the paper, we will not separately name semi-structured texts where it is not relevant, but will instead understand structured texts to subsume semi-structured texts.

It is known that LLMs can handle structured inputs, having encountered the common formalisms during their basic training. Many studies explore how effectively LLMs can create structured documents from natural language text.
In contrast, this paper focuses on the ability of LLMs to process already structured texts. We do not aim to convert natural language descriptions of, e.g., graphs or processes, into  representations structured according to some formalism. Rather, we investigate how well LLMs can process or restructure inputs that are already structured.

Restructuring structured documents has practical applications, particularly in writing documents that include formatting and layout information, such as Markdown, HTML, or LaTeX. By inserting or adjusting such formatting, LLMs can support authoring activities beyond merely generating new content. Further applications include converting between different document formats, which is essential when data needs to be reformatted for automatic processing.
In software development, the capabilities of an LLM can replace traditionally programmed conversion routines, which are often expensive to develop and test. Integrating an LLM can reduce software development costs and enable more flexible and powerful solutions than would be achievable with classical programming. However, this approach incurs ongoing operational costs if a paid LLM-as-a-Service is utilized.

Although the tasks performed by the LLM in editing structured documents may seem less demanding than other currently researched tasks, they can still bring significant labor savings and efficiency gains. The prerequisite for this to be useful is that the application of the LLM for these tasks incurs little effort. 
This paper addresses the following research question:

(RQ) Can LLMs be applied for editing structured or semi-structured documents with little effort?

By 'little effort,' we mean that simple, quickly designed prompts should suffice, and the outputs of the LLM should be of high quality, requiring minimal manual post-processing.
'Editing semi-structured documents' refers to modifying their structure rather than their semantic content. 
To our knowledge, this question has not yet been investigated in research. 

\section{Related work} \label{sec:relwork}
This study offers a qualitative exploration of an LLM's ability to transform structured inputs or convert structured inputs from one format to another. No previous work explicitly investigating this topic was identified. 

The most closely related work focuses on LLM table understanding. For example, Singha et al.\ \cite{singhaTabularRepresentationNoisy2023} and Sui et al.\ \cite{suiTableMeetsLLM2024} conduct benchmark tests to evaluate LLM performance in interpreting structural tables. These studies present tables in various formats, including HTML, JSON, or Markdown to a range of LLMs, which then answer questions about the table data or table structure in natural language. These tests are conducted on a large scale, with performance assessed automatically.
We also reviewed several applications of LLMs that operate on or produce structured outputs similar to those investigated here, as summarized in \cref{tab:relatedworks}. However, an extensive literature review of such applications is beyond the scope of this paper.

\begin{table}[tbh]
\begin{tabularx}{\textwidth}{@{}c X X X@{}}
\hline
Ref & Purpose & Input & Output \\
\hline
\cite{wuCorefQACoreferenceResolution2020} & Identify co-reference & NL & SEM (XML) \\
\cite{wuCorefQACoreferenceResolution2020} & Resolve co-reference & SEM (XML) & STRUC \\
\cite{aroraLanguageModelsEnable2023} & Extract data & SEM (HTML, TXT, XML) & STRUC (DB) \\
\cite{polakExtractingAccurateMaterials2024} & Extract data & NL (scientific articles) & STRUC (DB) \\
\cite{minRecentAdvancesNatural2023} & NLP tasks & NL & Diverse (SEM, NL, etc.) \\
\cite{yeLanguageAllGraph2024} & Perform graph tasks & SEM (graph) & NL (e.g., a category) \\
\cite{chenLabelfreeNodeClassification2024} & Create training data & STRUC (graph) & STRUC (training data) \\
\cite{jiangStructGPTGeneralFramework2023} & Answer questions & STRUC (Diverse) & NL or STRUC (queries) \\
\cite{fillConceptualModelingLarge2023} & Draw diagrams & NL & STRUC (JSON diagrams) \\
\cite{helfrich-schkarbanenkoMathematikUndChatGPT2023} & Create math exercises & NL & STRUC (LaTeX math) \\
\cite{helfrich-schkarbanenkoMathematikUndChatGPT2023} & Phrase math formulae & STRUC (LaTeX math) & NL \\
\cite{helfrich-schkarbanenkoMathematikUndChatGPT2023} & Create drawings & NL & STRUC (TikZ code) \\
\cite{xiaFOFOBenchmarkEvaluate2024} & Create documents & NL + STRUC (an example) & STRUC (like the example) \\
\cite{labanChatExecutableVerifiable2023} & Edit (not create) texts & NL & NL \\
\hline
exp1 & Edit (not create) docs & SEM (LaTeX) & SEM (LaTeX) \\
exp2 & Edit (not create) docs & STRUC (RIS) & STRUC (OPUS XML) \\
\hline
\end{tabularx}
\caption{Applications processing structured (STRUC) or semi-structured (SEM) texts. NL indicates natural language, DB indicates database entries. 'Create' means 'generate new content'. exp1 and exp2 refer to the case studies conducted in this paper.}
\label{tab:relatedworks}
\end{table}

Wu et al.\ \cite{wuCorefQACoreferenceResolution2020} present an application for  co-reference resolution, a common task in Natural Language Processing (NLP). Their application  queries an LLM twice. The first query tags a natural language input with XML tags, while the second query consumes  this semi-structured result as input and yields a structured  output.
Two further applications use LLMs for extracting  data  into a queryable, highly structured tabular format.  One  processes various types of semi-structured documents (e.g., HTML, TXT, XML) \cite{aroraLanguageModelsEnable2023}, while the second scans  scientific articles, i.e., natural language texts, to retrieve  cooling rates of metallic glasses  \cite{polakExtractingAccurateMaterials2024}.
An extensive overview of applications of LLMs for tasks encountered in NLP reports works where  LLMs produce structured outputs from unstructured inputs \cite{minRecentAdvancesNatural2023}.
Several papers focus on processing graphs with LLMs. One study describes the geometric structure of graphs in natural language and then utilizes the LLM to perform graph tasks, specifically node classification  \cite{yeLanguageAllGraph2024}.  In \cite{chenLabelfreeNodeClassification2024}, an LLM is employed for generating structured training data to train a Graph Neural Network for node classification, thus avoiding the high costs of using the LLM for node classification directly.
Jiang et al.\ \cite{jiangStructGPTGeneralFramework2023} present StructGPT,  a system that interfaces with various structured data pools, specifically, databases and knowledge graphs. StructGPT retrieves data from the pools and passes it to an LLM, which is tasked to answer questions based on this structured data. The LLM either provides the answer directly or generates a database query that can retrieve the answer.

The capability of ChatGPT-4 to generate entity-relationship diagrams, business process models in BPMN, and UML class diagrams from  descriptions phrased in natural language is evaluated in \cite{fillConceptualModelingLarge2023}. The models and diagrams are generated using representations based on JSON. 
In \cite{helfrich-schkarbanenkoMathematikUndChatGPT2023}, LaTeX is proposed as a means to communicate mathematical concepts and create drawings with an LLM. The LLM is tasked to generate mathematical exercises and corresponding solutions in LaTeX. It is also applied to translate LaTeX formulas into natural language, which can be read aloud to visually impaired persons. Furthermore, the LLM is tasked with creating drawings using TikZ commands, a language for producing vector graphics in LaTeX documents. Xia et al.\ 
\cite{xiaFOFOBenchmarkEvaluate2024} contribute a benchmark dataset designed  to evaluate the capabilities of LLMs in producing structured outputs across a range of application domains and document formats. Their benchmark dataset comprises prompts which instruct an LLM to create a document in a specific format with the format specified by an example. 
A further LLM  is applied to assess whether the evaluated  LLMs successfully generated documents in the required format. Laban et al.\ 
\cite{labanChatExecutableVerifiable2023} employ LLMs for editing (not generating) unstructured natural language texts. Their system aims to assist authors in writing.
While these studies involve LLMs processing structured inputs or producing structured outputs,  none of them investigates the capability of LLMs for  reformatting or restructuring structured documents.

\section{Method}
To address the research question, we conduct experiments using various document formats. The research adopts a  qualitative rather than a quantitative approach. The number of experiments is deliberately kept low,  and the results are reviewed and  evaluated "by hand". 
This approach allows for identifying details and making unexpected observations that automated tests with large datasets might overlook, as they typically provide only percentages of correctness as, e.g., in \cite{xiaFOFOBenchmarkEvaluate2024}.
The research aims to investigate tasks that closely resemble real-world scenarios. 
We conducted two series of experiments. The first series involves documents formatted with LaTeX, a widely-used typesetting language familiar to ChatGPT. 
The second series uses less common document formats. Here, ChatGPT is tasked with converting RIS records into an XML format used by OPUS. RIS is a standardized markup format for exchanging bibliographic information between literature management programs, while OPUS is a software used by institutions to set up publication databases \cite{kooperativebibliotheksverbundberlin-brandenburgkobvWasIstOPUS2023}. 
To ensure meaningful insights and avoid introducing unintentional biases, we use realistic sample documents.
In the first series of experiments, a LaTeX-formatted table taken from a research paper \cite{weberLargeLanguageModels2024} was processed. The highly specific technical terms originally presented in this table were replaced with more neutral terms using ChatGPT, without altering the structure of the table.
Example documents for the second series are obtained from real university servers. 

Experiments were conducted using ChatGPT (then based on GPT-3.5) through OpenAI's chat interface on April 29 and May 1, 2024. Each experiment's prompt was input into the interface, and the model's response was then analyzed externally. Chat history was cleared after each experiment to ensure independent processing.
The input documents, prompts, and outputs are available online in an electronic appendix \cite{weberWeberi2024_AKWI_structured_gpt_experiments2024}.

\section{Experiment Series 1: Restructuring and reformatting LaTeX}\label{sec:latexexp}

\subsection{Sample Data and Prompts}
\begin{figure*}[h!]
    \centering  
  \includegraphics[width=0.7\textwidth]{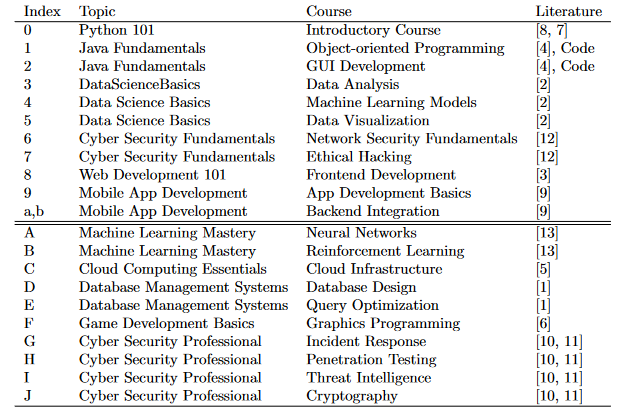}    \\
  \medskip

  \includegraphics[width=\textwidth]{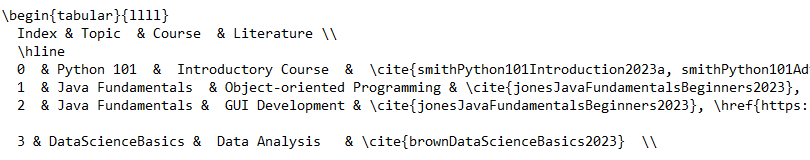}   
  \caption{Sample LaTeX table used for experiments}\label{fig:exptab0}   
  \end{figure*}

This experiment series comprises four steps in which the LaTeX table is progressively edited. The prompt for each step consists of an instruction and  a table in LaTeX format, with the chat history cleared after each LLM query. \Cref{tab:prompts-series1} lists the prompts.
\Cref{fig:exptab0} depicts the table and  a piece of its LaTeX definition before the first edit. 

\begin{table*}[h!]
    \footnotesize
\begin{tabularx}{\textwidth}{lXl}
      \hline
        No & Prompt  & Result\\
        \hline
        1 & \texttt{I will give you a LaTeX table. Please delete the first colum. ''' }   & \cref{fig:exptab1} \\
        2a & \texttt{I will give you a LaTeX table. Please swap the two last columns. '''    } & \\
        2b & \texttt{I will give you a LaTeX table. Please swap the "{}Course"{}  and "{}Literature"{}  columns'''    }   & \cref{fig:exptab2b} \\
        3a & \texttt{I will give you a LaTeX table. I want you to reduce the number of lines as follows. Some lines only differ in the last column. Please collapse these lines in one line. Collect their last colum data.    } & \cref{fig:exptab3a}\\
        3b & identical to 3a \\
        4a & \texttt{I will give you a LaTeX table. Please format the entries in the "{}Course"{}  column in Italics. Keep the formatting of separating commas as it is. '''   } & \\
        4b & \texttt{I will give you a LaTeX table. Please format the entries in the "{}Courses"{} column in Italics. There may be multiple entries in one cell, separated by commas. Keep the formatting of separating commas as it is. '''  }& \\
        4c & \texttt{I will give you a LaTeX table. Please format the entries in the "{}Courses"{}  column in Italics. There may be multiple entries in one cell, separated by commas. Spare the commas out. ''' } &   \cref{fig:exptab4c} \\
        4d & \texttt{I will give you a LaTeX table. Please format the entries in the "{}Course"{}  column in Italics excluding the commas. '''   }  &\\
        \hline
\end{tabularx}
\caption{Prompts for LaTeX restructuring experiments. Complete prompts and results are available in \cite{weberWeberi2024_AKWI_structured_gpt_experiments2024}.}\label{tab:prompts-series1}
\end{table*}

To show the generated LaTeX tables and test the generated  LaTeX commands, we manually inserted the LLM-generated tables into LaTeX documents such that a PDF could be created. The resulting tables are depicted in \cref{fig:exptab1,fig:exptab2b,fig:exptab3a,fig:exptab4c}. 
Protocols of the experiments, along with prompts and complete versions of the input and output LaTeX tables, can be found in the electronic appendix  \cite{weberWeberi2024_AKWI_structured_gpt_experiments2024}.

\subsection{Results}

In all experiments, ChatGPT generated tables in correct LaTeX syntax that the LaTeX compiler processed without issues. It was able to make all desired changes, although in some experiments, this was achieved only after modifying the prompts, as reported below. The results were not consistently reproducible, meaning that identical queries with cleared chat history sometimes, but not always, produced different outputs. This variability might stem from ChatGPT's temperature settings.

\begin{figure*}[h!]
    \centering  
  \includegraphics[width=0.7\textwidth]{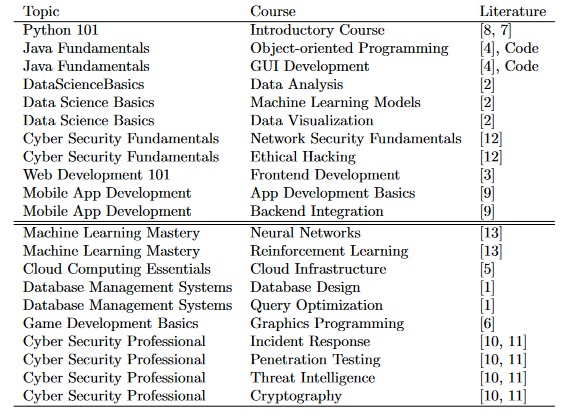}    
  \caption{LaTeX table generated by Prompt 1}\label{fig:exptab1}   
  \end{figure*}

\begin{figure*}[h!]
    \centering  
  \includegraphics[width=0.8\textwidth]{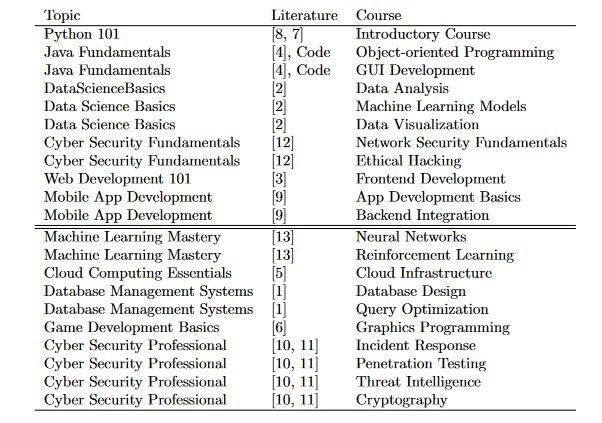}    
  \caption{LaTeX table generated by Prompt 2b}\label{fig:exptab2b}   
  \end{figure*}

\begin{figure*}[h!]
    \centering  
  \includegraphics[width=0.9\textwidth]{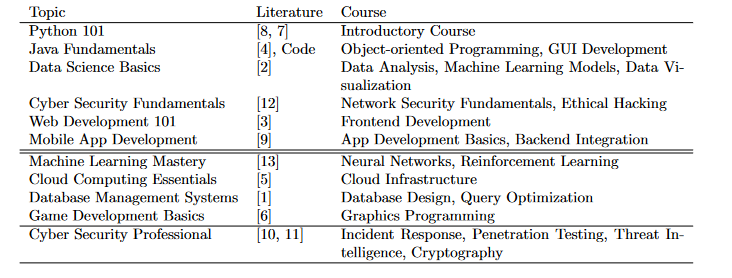}    
  \caption{LaTeX table generated by Prompt 3a}\label{fig:exptab3a}   
  \end{figure*}

Prompt 1 produced the desired result, see \cref{fig:exptab1}.
Prompt 2a returned the input table nearly unchanged with only a subtle modification in one cell: the content of the last column of the third row  ("{}[8, 7]{}") were replaced by the content of the cell above it ("{}[4], Code"{}). 
Prompt 2b produced the desired result, as shown in the \cref{fig:exptab2b}.
Prompt 3a successfully restructured the table as requested. It merged rows 3 to 5, despite differences in the spelling of the "{}Topic"{} column, and adopted the spelling "{}Data Science Basics"{}, as illustrated in \cref{fig:exptab3a}. Additionally, it added a dividing line before the last table row. A second query with an unchanged prompt 3b also correctly restructured the table, but this time it adopted the spelling "{}DataScienceBasics"{} when merging rows 3 to 5 and did not generate an additional dividing line.

  \begin{figure*}[h!]
    \centering  
  \includegraphics[width=0.9\textwidth]{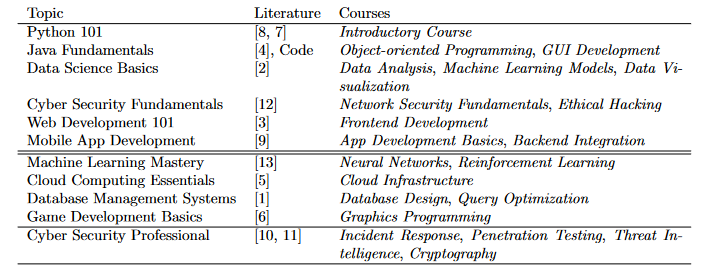}    
  \caption{LaTeX table generated by Prompt 4c}\label{fig:exptab4c}   
  \end{figure*}

In step 4, ChatGPT was instructed to format specific table contents using various prompt variants, as shown \cref{tab:prompts-series1}. Specifically, certain table cells' texts were to be printed in italics, excluding commas.
In all queries, the specified table contents were reformatted in Italics. However, ChatGPT  only succeeded in  skipping the commas as requested in some queries. Repeated queries with identical prompts sometimes succeeded and sometimes failed. The result of a successful query using Prompt 4c is depicted in \cref{fig:exptab4c}.
In some step 4 queries, ChatGPT added  extra LaTeX commands. Specifically, it embedded the provided LaTeX text fragment  in a LaTeX table environment (a structure that allows controlling the placement of the table and the inclusion of a caption and label) or even provided a complete LaTeX document (excluding the bibliography).

\section{Experiment Series 2: Converting structured documents} \label{sec:ris-xmlexp}

\subsection{Sample Data and Prompts}
The second series of experiments  investigates the capability of the LLM in converting  structured documents between different formats.    
We use RIS and OPUS XML data originating from the OPUS servers of  Landshut University of Applied Sciences\footnote{\url{https://opus4.kobv.de/opus4-haw-landshut}}  (HAWL) and  Technical University Rosenheim\footnote{\url{https://opus4.kobv.de/opus4-rosenheim}} (THR) for the case study. Both servers offer the option to export stored publications in RIS and in XML format. 
\cref{tab:exp2} gives an overview over the data used for the experiments. The differing numbers of fields show that the RIS and XML exports are not as uniform as might be expected.

\begin{table}[h]
    \centering

    \begin{tabular}{llllccc}
    \hline
   Ref. & Id & Source & Conf. & 1-shot example & RIS  & XML \\
    \hline
   \cite{seehuberEtherCATGatewayFur2022} &  \textsc{Seehuber} & HAWL & 3. Symp ESI & X & 17 & 38 \\
   \cite{munchIntegrationSecurityGateway2022} &    \textsc{Muench} & HAWL  & 3. Symp ESI & & 16 & \textit{35} \\
   \cite{ZugschwertGoeschlIbanezetal.2021} & \textsc{Zugschwert} & HAWL  & - na - & & 17 & \textit{23} \\
   \cite{seligerHighfrequencyPerformanceDegradation2024} &  \textsc{Seliger} & THR &  CIPS 2024 & & 18 & \textit{36} \\
    \hline
    \end{tabular}
    \caption{ Data for series 2 of experiments. The table shows the number of fields in the RIS exports, the number of fields in the OPUS export Seehuber.xml, which serves as an example, and the number of XML fields generated by the LLM (printed in italics). }
    \label{tab:exp2}
    \end{table}
    
\begin{figure*}[hbt]
        \centering  
      \includegraphics[width=1\textwidth]{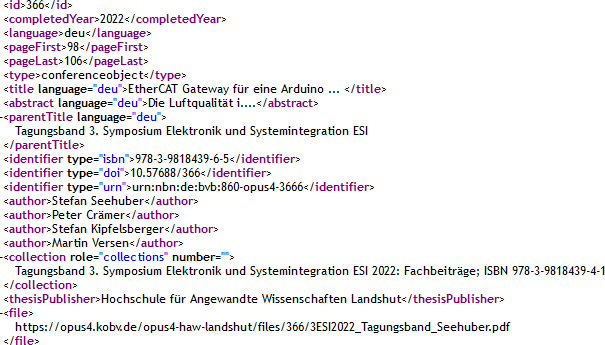}    
      \caption{Excerpt of \textsc{Seehuber.xml} as exported from the HAWL OPUS server}\label{fig:exseel}   
      \end{figure*}

    All publications were exported in RIS format and \textsc{Seehuber} also in XML format. \textsc{Seehuber.ris} and \textsc{Seehuber.xml} serve as  one-shot prompt examples  for ChatGPT.  \cref{fig:seeexp} shows  \textsc{Seehuber.ris}, and \cref{fig:exseel} illustrates some fields of  \textsc{Seehuber.xml}.
    \textsc{Seehuber.ris} and \textsc{Muench.ris} are conference contributions to the same conference and therefore contain several  identical fields. \textsc{Zugschwert.ris} and \textsc{Seliger.ris}, which are also conference contributions, do not contain all fields in \textsc{Seehuber.ris}, but they do contain additional fields not present in \textsc{Seehuber.ris}. 
 
    \begin{figure}[thb]
        \footnotesize
    \begin{verbatim}
        TY  - CONF
        A1  - Seehuber, Stefan
        A1  - Crämer, Peter
        A1  - Kipfelsberger, Stefan
        A1  - Versen, Martin
        A2  - Artem, Ivanov
        A2  - Marc, Bicker
        A2  - Peter, Patzelt
        T1  - EtherCAT Gateway für eine [...] Visualisierung [...]
        T2  - Tagungsband 3. Symposium Elektronik und Systemintegration ESI
        N2  - Die [...]
        Y1  - 2022
        UR  - https://opus4.kobv.de/opus4-haw-landshut/frontdoor/index/index/docId/366
        UR  - https://nbn-resolving.org/urn:nbn:de:bvb:860-opus4-3666
        SN  - 978-3-9818439-6-5
        SP  - 98
        EP  - 106
        ER  - 
    \end{verbatim}    
\caption{\textsc{Seehuber.ris} as exported from the HAWL OPUS server}\label{fig:seeexp}
    \end{figure}

    The prompt for ChatGPT is constructed using  a  one-shot pattern. It includes a brief instruction, \textsc{Seehuber.ris} and \textsc{Seehuber.xml} at positions  
    \texttt{\%\%1}  and   \texttt{\%\%2}, and one of  \textsc{Muench.ris}, \textsc{Zugschwert.ris} and \textsc{Seliger.ris} as a task at position \texttt{\%\%3}:

    {\footnotesize
 \begin{verbatim}
I will input a ris-document. Please convert it to Opus-XML.  First, you
 will be provided with an example input and output. 
Here is the example input:  '''  %%1  ''' 
Here is the example output: '''  %%2  '''   
Here is the ris-document you must convert:'''  %%3  '''  
\end{verbatim}
}

Consequently, for each experiment, the prompt contained  \textsc{Seehuber.ris} and \textsc{Seehuber.xml} as the example, and the RIS of one publication. ChatGPT was queried once with each of the resulting prompts. The XML it generated was then compared to the original corresponding  RIS and to the example \textsc{Seehuber.xml}. 

\subsection{Additional prompts}

ChatGPT was also prompted to convert a RIS into OPUS XML with a zero-shot prompt, i.e., a prompt lacking an example. The
zero-shot prompting yielded a syntactically correct XML with a plausible structure and plausibly named fields, but differing from an actual  OPUS XML export. This indicates that ChatGPT did not learn these formats or their interconnections during its training.
ChatGPT was also asked about details of the publication by Seehuber et al.\ \cite{seehuberEtherCATGatewayFur2022} and by Zugschwert et al.\  \cite{ZugschwertGoeschlIbanezetal.2021} and stated  not to know them as follows: \texttt{
  I don't have access to specific publications or writings  by SeeHuber, 
  Crämer, and Kippelsberger in 2022 regarding Luftqualität (air quality).
  […] my last update in January 2022.}

\subsection{Results}\label{sec:xml-results}

\begin{figure*}[htb]
    \centering  
  \includegraphics[width=1\textwidth]{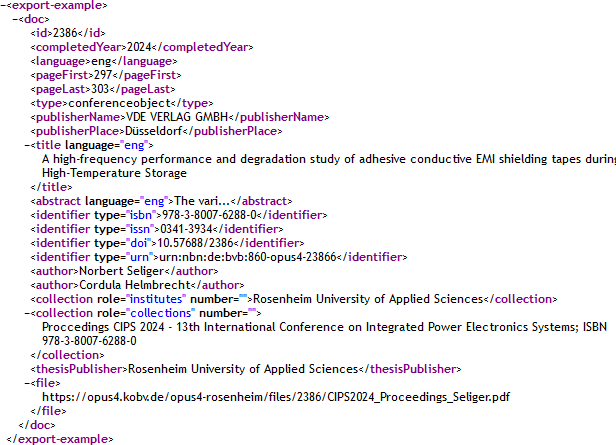}    
  \caption{ Excerpt from \textsc{Seliger.xml} as generated by ChatGPT}\label{fig:exseliger}   
  \end{figure*}

 ChatGPT  generated the XML format for all prompts without any syntactic errors.  \cref{fig:exseliger} shows excerpts of the output generated for \textsc{Seliger.ris}. The complete outputs of all experiments can be found in the electronic appendix \cite{weberWeberi2024_AKWI_structured_gpt_experiments2024}.
 ChatGPT correctly created all XML fields present in the example XML.
   For author fields occurring in varying numbers, it created the correct number of fields in the XML and filled them correctly with the authors' names as values.
  The names occurring  in the format "{}Lastname, Firstname{}"\ in RIS documents were transferred to XML as "{}Firstname Lastname{}"{} matching the provided example.
  The RIS files do not contain language information. ChatGPT added this information to match the actual language of the publication, replacing "{}deu{}"{} with "{}eng{}"{}, for example, \verb|<title language="deu">|  $\rightarrow$   \verb|<title language="eng">|, according to the provided XML example.
 Fields that were present in the example \textsc{Seehuber.xml} but not in the example \textsc{Seehuber.ris} were correctly filled in the generated XML documents; e.g., \newline
 \verb|PU VDE VERLAG GMBH| $\rightarrow$ \verb|<publisherName>VDE VERLAG GMBH</publisherName>| \newline
 \verb|CY Düsseldorf| $\rightarrow$ \verb|<publisherPlace>Düsseldorf</publisherPlace>|

RIS fields of type \texttt{KW} (keywords) that were present in the new RIS documents, e.g., in \textsc{Seliger.ris}, but not in the example \textsc{Seehuber.ris} and \textsc{Seehuber.xml}, were not added to the generated XML, i.e., \textsc{Seliger.xml}, meaning that no new field identifiers for keywords  were hallucinated.
For RIS fields of type \texttt{A2} (editors of conference proceedings) that were not provided in the \textsc{Zugschwert.ris} or the \textsc{Seliger.ris}, ChatGPT did not generate entries in the XML, hence editor names were neither copied nor hallucinated.

\begin{table}[p] 
\footnotesize 
\begin{tabularx}{\textwidth}{l X}
 \multicolumn{2}{l}{Values from \textsc{Seehuber.ris} and \textsc{Seehuber.xml} } \\
\hline
\verb|Y1|& \textcolor{Sepia}{2022}     \qquad \qquad   \verb|SN| \quad   978-3-9818439-6-5    \\
\verb|A1| &  \textcolor{red}{Seehuber}, Stefan   \\
\verb|A1| &  Crämer, Peter   \\
\verb|A1| &   Kipfelsberger, Stefan \\
\verb|A1| &   Versen, Martin \\
\verb|T2|    & \textcolor{OliveGreen}{Tagungsband} \textcolor{blue}{3}. Symposium Elektronik und Systemintegration \textcolor{blue}{ESI}                                         \\       
\verb|UR|      &\textcolor{RedViolet}{https://opus4.kobv.de/opus4-haw-landshut}/frontdoor/index/index/docId/\textcolor{MidnightBlue}{366 }      \\
c& {\textcolor{OliveGreen}{Tagungsband} 3. Symposium   Elektronik und Systemintegration ESI \textcolor{Sepia}{2022}: Fachbeiträge; ISBN   978-3-9818439-4-1}      \\
f& {\textcolor{RedViolet}{https://opus4.kobv.de/opus4-haw-landshut}/files/\textcolor{MidnightBlue}{366}/\textcolor{blue}{3ESI}\textcolor{Sepia}{2022}\_\textcolor{OliveGreen}{Tagungsband}\_\textcolor{red}{Seehuber}.pdf  } \\
\hline
\\[0.05ex]
\multicolumn{2}{l}{Values from \textsc{Muench.ris} and generated by ChatGPT}  \\
\hline
\verb|Y1| & \textcolor{Sepia}{2022}    \qquad \qquad  \verb|SN| \quad   978-3-9818439-6-5    \\ 
\verb|A1| &  \textcolor{red}{Münch}, Andreas  \\
\verb|A1| &   Frauenschläger, Tobias \\
\verb|A1| &  Mottok, Jürgen \\
\verb|T2|         &\textcolor{OliveGreen}{Tagungsband} \textcolor{blue}{3}. Symposium Elektronik und Systemintegration \textcolor{blue}{ESI}                                \\
\verb|UR|        &\textcolor{RedViolet}{https://opus4.kobv.de/opus4-haw-landshut}/frontdoor/index/index/docId/\textcolor{MidnightBlue}{365}               \\ 
\textit{c}& \textit{\textcolor{OliveGreen}{Tagungsband} 3. Symposium   Elektronik und Systemintegration ESI \textcolor{Sepia}{2022}: Fachbeiträge; ISBN   978-3-9818439-4-1}              \\
\textit{f}&\textit{\textcolor{RedViolet}{https://opus4.kobv.de/opus4-haw-landshut}/files/\textcolor{MidnightBlue}{365}/\textcolor{blue}{3ESI}\textcolor{Sepia}{2022}\_\textcolor{OliveGreen}{Tagungsband}\_\textcolor{red}{Münch}.pdf}                           \\
\hline
\\[0.05ex]
\multicolumn{2}{l}{Values from \textsc{Zugschwert.ris} and generated by ChatGPT}  \\ 
\hline
\verb|Y1|& \textcolor{Sepia}{2021}    \qquad  \qquad  \verb|SN| \quad   - na-     \\ 
\verb|A1| &  \textcolor{red}{Zugschwert}, Christina  \\
\verb|A1| &   Göschl, Sebastian \\
\verb|A1| &  Ibanez, Federico Martin \ \\
\verb|A1| &   Pettinger, Karl-Heinz \\
\verb|T2|       & - na -                     \\
\verb|UR|        &\textcolor{RedViolet}{https://opus4.kobv.de/opus4-haw-landshut}/frontdoor/index/index/docId/\textcolor{MidnightBlue}{303}                              \\
\textit{c}&\textit{\textcolor{OliveGreen}{Tagungsband} 3. Symposium   Elektronik und Systemintegration ESI \textcolor{Sepia}{2021}: Fachbeiträge; ISBN   978-3-9818439-4-1}              \\
\textit{f}&\textit{\textcolor{RedViolet}{https://opus4.kobv.de/opus4-haw-landshut}/files/\textcolor{MidnightBlue}{303}/\textcolor{blue}{3ESI}\textcolor{Sepia}{2021}\_\textcolor{OliveGreen}{Tagungsband}\_\textcolor{red}{Zugschwert}.pdf }                     \\
\hline
\\[0.05ex]
\multicolumn{2}{l}{Values from \textsc{Seliger.ris} and generated by ChatGPT}  \\ 
\hline
\verb|Y1| & \textcolor{Sepia}{2024}     \qquad \qquad  \verb|SN| \quad   978-3-8007-6288-0     \\
\verb|A1| &  \textcolor{red}{Seliger}, Norbert   \\
\verb|A1| &   Helmbrecht, Cordula\\
\verb|T2|         &  \textcolor{OliveGreen}{Proccedings} \textcolor{blue}{CIPS} \textcolor{Sepia}{2024} - 13th International Conference on   Integrated Power Electronics Systems              \\
\verb|UR|         &\textcolor{RedViolet}{https://opus4.kobv.de/opus4-rosenheim}/frontdoor/index/index/docId/\textcolor{MidnightBlue}{2386}                                      \\
\textit{c}&\textit{ \textcolor{OliveGreen}{Proccedings} \textcolor{blue}{CIPS} \textcolor{Sepia}{2024} -   13th International Conference on Integrated Power Electronics Systems; ISBN   978-3-8007-6288-0} \\
\textit{f}&\textit{\textcolor{RedViolet}{https://opus4.kobv.de/opus4-rosenheim}/files/\textcolor{MidnightBlue}{2386}/\textcolor{blue}{CIPS}\textcolor{Sepia}{2024}\_\textcolor{OliveGreen}{Proceedings}\_\textcolor{red}{Seliger}.pdf}                  \\
\hline
 \end{tabularx}
\caption{Constructing strings.  \texttt{Y1},  \texttt{SN},  \texttt{T2},  \texttt{UR} are RIS tags and values. \textit{c} and \textit{f} signify the \texttt{<collection>} and the \texttt{<field>} XML fields. \textsc{Seehuber.xml} fields are part of the example XML included in the prompt. The remaining \texttt{c} and \texttt{f} values are generated by the LLM (printed in italics).}  \label{tab:patmatch}
\end{table}

For fields not present in the example \textsc{Seehuber.ris} but having more or less matching fields in the example \textsc{Seehuber.xml}, ChatGPT constructed appropriate entries in the XML documents. In constructing these values, ChatGPT worked in a very detailed manner. We analyze three occurrences more closely: 
First,  ChatGPT correctly derived document IDs, presumably by extracting them from the URLs  provided in the RIS tag \verb|UR|. Notably, RIS documents do not comprise a tag for the document ID,  while the XML format comprises a dedicated \verb|<id>| field (compare \cref{fig:seeexp} and \cref{fig:exseel}). 
Second, ChatGPT constructed the string value for the \verb|<collection role="collections">| XML field, presumably from the values provided in the   \verb|A1|, \verb|T2|,  \verb|Y1| and  \verb|SN| RIS tags, where the last part of this string and the \verb|SN| value do not completely match in the \textsc{Seehuber} example provided to ChatGPT. It should be noted that the RIS documents contain multiple  \verb|A1| fields (the authors), and ChatGPT consistently selected the value of the first  \verb|A1| in all experiments.
Third, it constructed the download links for the \verb|<file>| XML fields from multiple parts taken from several RIS fields and the document ID, which it had to extract from the \verb|UR|. It also  adapted the year \verb|2021| in the download link in  \textsc{Zugschwert.xml}. \cref{tab:patmatch}  juxtaposes the RIS fields containing the provided information and the constructed strings for each publication.

It should be emphasized that ChatGPT constructed the strings in the \verb|f| and \verb|c| fields from scratch, and that these strings were embedded within the complete XML by a single query to the LLM, rather than being constructed and positioned separately.
Not all the constructed values are  "{}correct{}"{} in the real world. While the derived document IDs are valid, the constructed links are not. The link in \textsc{Muench.xml} is a near miss, as the actual link only differs in the spelling   "{}Muench{}"{} versus  "{}Münch{}"{}   from the generated one. 

\section{Discussion and Conclusion}
This paper contributes a qualitative investigation on the original research question whether
an LLM can successfully edit semi-structured documents and transform structured documents when prompted with basic and straightforward instructions. We conducted two case studies comprising multiple experiments, one restructuring a LaTeX table, and one converting RIS documents to OPUS XML format.
Our results  indicate that the research question has a  positive answer.
In all experiments, the LLM produced syntactically correct documents which could be further processed without issues.
The research followed a qualitative approach,  conducting a limited number of experiments. Additional and broader experiments are needed to determine if this finding generalizes to other restructuring tasks and LLMs.

While  related studies on LLM capabilities \cite{zhouInstructionFollowingEvaluationLarge2023a, xiaFOFOBenchmarkEvaluate2024, singhaTabularRepresentationNoisy2023, suiTableMeetsLLM2024} conduct massive tests and  evaluate results automatically (e.g., through LLMs \cite{xiaFOFOBenchmarkEvaluate2024}), our qualitative approach includes a  comprehensive and in-depth manual analysis of the results. This allows us to contribute the following detailed observations.
In the LaTeX experiments described in \cref{sec:latexexp}, the LLM was tasked with restructuring a LaTeX table. We found that  the LLM understood concepts related to tables such as "{}row"{}, "{}column"{} and "{}cell"{} very well. Referring to table columns by their titles worked better than referencing them by their position (i.e., "{}last"{}).  
While the LLM reliably recognized and handled the structure explicated by LaTeX annotations, it 
struggled with recognizing commas as structure indicators.
These observations lead to the hypothesis that explicit structural annotations (such as LaTeX commands) may enhance  an LLM's understanding of tasks and data provided in prompts, thereby yielding better outputs. Specifically, they might improve the LLM's instruction-following and format-following capabilities  \cite{zhouInstructionFollowingEvaluationLarge2023a, xiaFOFOBenchmarkEvaluate2024}, which are crucial when developing LLM-integrated applications \cite{weberLargeLanguageModels2024}.  Further experiments  exploring this hypothesis  will be valuable.

The RIS$\rightarrow$XML experiments in \cref{sec:ris-xmlexp} reveal that the LLM has impressive pattern matching skills, which become evident in the strings it constructed, compare \cref{tab:patmatch} . 
 It seems plausible that its working principle involves identifying relationships (i.e., patterns) between RIS  and XML elements  in the example documents and replicate these in the documents it was tasked with generating.
Some data elements generated are correct with respect to the real world, while other data elements are near misses or completely deviating, as elaborated in \cref{sec:xml-results}. However, it does not seem appropriate to label the latter data elements as   "{}hallucinated{}", as the process that generated them is comprehensible and reasonable, albeit overgeneralizing to some extent. This pattern matching behavior deserves further investigation, as it may constitute a novel approach to understanding the processes leading to hallucinations in LLMs \cite{jiSurveyHallucinationNatural2023}.

\bibliographystyle{alpha}

\begin{thebibliography}{MFM22x}

\bibitem[Ar23]{aroraLanguageModelsEnable2023}
Arora, S.; Yang, B.; Eyuboglu, S.; Narayan, A.; Hojel, A.; Trummer, I.; Ré, C.:
Language Models Enable Simple Systems for Generating Structured Views of Heterogeneous Data Lakes. 2023, doi: \url{http://arxiv.org/abs/2304.09433}.

\bibitem[Ch24]{chenLabelfreeNodeClassification2024}
Chen, Z.; Mao, H.; Wen, H.; Han, H.; Jin, W.; Zhang, H.; Liu, H.; Tang, J.:
Label-Free Node Classification on Graphs with Large Language Models (LLMS). 2024, doi: \url{http://arxiv.org/abs/2310.04668}.

\bibitem[FFK23]{fillConceptualModelingLarge2023}
Fill, H.-G.; Fettke, P.; Köpke, J.:
Conceptual Modeling and Large Language Models: Impressions From First Experiments With ChatGPT. Enterprise Modelling and Information Systems Architectures (EMISAJ) 18, pp. 1–15, 2023, doi: \url{https://doi.org/10.18417/emisa.18.3}.

\bibitem[He23]{helfrich-schkarbanenkoMathematikUndChatGPT2023}
Helfrich-Schkarbanenko, A.:
Mathematik und ChatGPT: Ein Rendezvous am Fuße der technologischen Singularität. Springer Berlin Heidelberg, Berlin, Heidelberg, 2023, isbn: 978-3-662-68209-8.

\bibitem[Ji23a]{jiSurveyHallucinationNatural2023}
Ji, Z.; Lee, N.; Frieske, R.; Yu, T.; Su, D.; Xu, Y.; Ishii, E.; Bang, Y. J.; Madotto, A.; Fung, P.:
Survey of Hallucination in Natural Language Generation. ACM Comput. Surv. 55 (12), 248:1–248:38, 2023, doi: \url{https://doi.org/10.1145/3571730}.

\bibitem[Ji23b]{jiangStructGPTGeneralFramework2023}
Jiang, J.; Zhou, K.; Dong, Z.; Ye, K.; Zhao, W. X.; Wen, J.-R.:
StructGPT: A General Framework for Large Language Model to Reason over Structured Data. 2023, doi: \url{http://arxiv.org/abs/2305.09645}.

\bibitem[Ko26]{kooperativebibliotheksverbundberlin-brandenburgkobvWasIstOPUS2023}
Kooperative Bibliotheksverbund Berlin-Brandenburg (KOBV):
Was Ist OPUS? - OPUS 4 Handbuch, \url{https://www.opus-repository.org/userdoc/introduction.html}, Freitag, 29. September 2023, 11:40:26, visited on: 04/29/2024.

\bibitem[La23]{labanChatExecutableVerifiable2023}
Laban, P.; Vig, J.; Hearst, M. A.; Xiong, C.; Wu, C.-S.:
Beyond the Chat: Executable and Verifiable Text-Editing with LLMs. 2023, doi: \url{http://arxiv.org/abs/2309.15337}.

\bibitem[MBZ13]{madaniSemistructuredDocumentsMining2013}
Madani, A.; Boussaid, O.; Zegour, D. E.:
Semi-Structured Documents Mining: A Review and Comparison. Procedia Computer Science 22, pp. 330–339, 2013, doi: \url{https://doi.org/10.1016/j.procs.2013.09.110}.

\bibitem[MFM22]{munchIntegrationSecurityGateway2022}
Münch, A.; Frauenschläger, T.; Mottok, J.:
Integration of a Security Gateway for Critical Infrastructure into Existing PKI Systems. In: Tagungsband 3. Symposium Elektronik Und Systemintegration ESI. Pp. 88–97, 2022, isbn: 978-3-9818439-6-5, doi: \url{https://doi.org/10.57688/365}.

\bibitem[Mi23]{minRecentAdvancesNatural2023}
Min, B.; Ross, H.; Sulem, E.; Veyseh, A. P. B.; Nguyen, T. H.; Sainz, O.; Agirre, E.; Heintz, I.; Roth, D.:
Recent Advances in Natural Language Processing via Large Pretrained Language Models: A Survey. ACM Computing Surveys 56 (2), 30:1–30:40, 2023, issn: 0360-0300, doi: \url{https://doi.org/10.1145/3605943}.

\bibitem[PM24]{polakExtractingAccurateMaterials2024}
Polak, M. P.; Morgan, D.:
Extracting Accurate Materials Data from Research Papers with Conversational Language Models and Prompt Engineering. Nature Communications 15 (1), p. 1569, 2024, issn: 2041-1723, doi: \url{https://doi.org/10.1038/s41467-024-45914-8}.

\bibitem[Se22]{seehuberEtherCATGatewayFur2022}
Seehuber, S.; Crämer, P.; Kipfelsberger, S.; Versen, M.:
EtherCAT Gateway für eine Arduino basierte LuftqualitätsMessung zur Visualisierung an eine Beckhoff SPS. In: Tagungsband 3. Symposium Elektronik und Systemintegration ESI. Pp. 98–106, 2022, doi: \url{https://doi.org/10.57688/366}.

\bibitem[SH24]{seligerHighfrequencyPerformanceDegradation2024}
Seliger, N.; Helmbrecht, C.:
A High-Frequency Performance and Degradation Study of Adhesive Conductive EMI Shielding Tapes during High-Temperature Storage. In: Proceedings CIPS 2024 - 13th International Conference on Integrated Power Electronics Systems. VDE VERLAG GMBH, pp. 297–303, 2024.

\bibitem[Si23]{singhaTabularRepresentationNoisy2023}
Singha, A.; Cambronero, J.; Gulwani, S.; Le, V.; Parnin, C.:
Tabular Representation, Noisy Operators, and Impacts on Table Structure Understanding Tasks in LLMs. 2023, doi: \url{http://arxiv.org/abs/2310.10358}.

\bibitem[Su24]{suiTableMeetsLLM2024}
Sui, Y.; Zhou, M.; Zhou, M.; Han, S.; Zhang, D.:
Table Meets LLM: Can Large Language Models Understand Structured Table Data? A Benchmark and Empirical Study. In: Proceedings of the 17th ACM International Conference on Web Search and Data Mining. WSDM ’24, Association for Computing Machinery, New York, NY, USA, pp. 645–654, 2024, doi: \url{https://doi.org/10.1145/3616855.3635752}.

\bibitem[We24a]{weberLargeLanguageModels2024}
Weber, I.:
Large Language Models as Software Components: A Taxonomy for LLM-Integrated Applications, 2024, doi: \url{http://arxiv.org/abs/2406.10300}.

\bibitem[We24b]{weberWeberi2024_AKWI_structured_gpt_experiments2024}
Weber, I.:
Weberi/2024\_AKWI\_structured\_gpt\_experiments, \url{https://github.com/weberi/2024_AKWI_structured_gpt_experiments}, 2024, visited on: 07/06/2024.

\bibitem[Wu20]{wuCorefQACoreferenceResolution2020}
Wu, W.; Wang, F.; Yuan, A.; Wu, F.; Li, J.:
CorefQA: Coreference Resolution as Query-based Span Prediction. In (Jurafsky, D.; Chai, J.; Schluter, N.; Tetreault, J., eds.): Proceedings of the 58th Annual Meeting of the Association for Computational Linguistics. Association for Computational Linguistics, Online, pp. 6953–6963, 2020, doi: \url{https://doi.org/10.18653/v1/2020.acl-main.622}.

\bibitem[Xi24]{xiaFOFOBenchmarkEvaluate2024}
Xia, C.; Xing, C.; Du, J.; Yang, X.; Feng, Y.; Xu, R.; Yin, W.; Xiong, C.:
FOFO: A Benchmark to Evaluate LLMs’ Format-Following Capability. 2024, doi: \url{http://arxiv.org/abs/2402.18667}.

\bibitem[Ye24]{yeLanguageAllGraph2024}
Ye, R.; Zhang, C.; Wang, R.; Xu, S.; Zhang, Y.:
Language Is All a Graph Needs. 2024, doi: \url{http://arxiv.org/abs/2308.07134}.

\bibitem[Zh23]{zhouInstructionFollowingEvaluationLarge2023a}
Zhou, J.; Lu, T.; Mishra, S.; Brahma, S.; Basu, S.; Luan, Y.; Zhou, D.; Hou, L.:
Instruction-Following Evaluation for Large Language Models. 2023, doi: \url{http://arxiv.org/abs/2311.07911}.

\bibitem[Zu21]{ZugschwertGoeschlIbanezetal.2021}
Zugschwert, C.; Göschl, S.; Ibanez, F. M.; Pettinger, K.-H.:
Development of a Multi-Timescale Method for Classifying Hybrid Energy Storage Systems in Grid Applications. In. Pp. 1–7, 2021, doi: \url{https://doi.org/10.57688/303}.




\end{thebibliography}

\end{document}